\pdfoutput=1

\documentclass[11pt]{article}
\usepackage[dvipsnames]{xcolor}
\usepackage{acl}

\usepackage{times}
\usepackage{latexsym}
\usepackage{setspace}
\usepackage[T1]{fontenc}

\usepackage[utf8]{inputenc}

\usepackage{microtype}

\usepackage[ruled,vlined]{algorithm2e}
\usepackage{graphicx}
\usepackage{hyperref}
\usepackage{enumerate}
\usepackage{booktabs}
\usepackage{enumitem}
\usepackage{bbding}
\usepackage{colortbl}
\usepackage{diagbox}
\usepackage{xcolor}
\usepackage{tcolorbox}

\usepackage{hyperref}
\definecolor{F7E0D5}{RGB}{245,240,255}
\colorlet{mycolor}{White!0!F7E0D5}

\newcommand{\tabincell}[2]{
    \begin{tabular}{@{}#1@{}}
        #2
    \end{tabular}
}

\title{Unbiased Math Word Problems Benchmark for Mitigating Solving Bias}

\author{Zhicheng Yang$^{1,2}$, Jinghui Qin$^2$, Jiaqi Chen$^{2,3}$ \and Xiaodan Liang$^{1,2}$\thanks{~~Corresponding author.} \\
$^1$Shenzhen Campus of Sun Yat-sen University\\
$^2$Sun Yat-sen University \\
$^3$Dark Matter AI Inc.\\
\texttt{\{yangzhch6,qinjingh\}@mail2.sysu.edu.cn} \\
\texttt{\{jadgechen,xdliang328\}@gmail.com} }

\begin{document}
\maketitle

\begin{abstract}


In this paper, we revisit the solving bias when evaluating models on current Math Word Problem (MWP) benchmarks. However, current solvers exist solving bias which consists of data bias and learning bias due to biased dataset and improper training strategy.
Our experiments verify MWP solvers are easy to be biased by the biased training datasets which do not cover diverse questions for each problem narrative of all MWPs, thus a solver can only learn shallow heuristics rather than deep semantics for understanding problems. 
Besides, an MWP can be naturally solved by multiple equivalent equations while current datasets take only one of the equivalent equations as ground truth, forcing the model to match the labeled ground truth and ignoring other equivalent equations.   
Here, we first introduce a novel MWP dataset named UnbiasedMWP which is constructed by varying the grounded expressions in our collected data and annotating them with corresponding multiple new questions manually. Then, to further mitigate learning bias, we propose a Dynamic Target Selection (DTS) Strategy to dynamically select more suitable target expressions according to the longest prefix match between the current model output and candidate equivalent equations which are obtained by applying commutative law during training. The results show that our UnbiasedMWP has significantly fewer biases than its original data and other datasets, posing a promising benchmark for fairly evaluating the solvers' reasoning skills rather than matching nearest neighbors. And the solvers trained with our DTS achieve higher accuracies on multiple MWP benchmarks. The source code is available at \href{https://github.com/yangzhch6/UnbiasedMWP}{https://github.com/yangzhch6/UnbiasedMWP}.

\end{abstract}

\section{Introduction}
Math Word Problem (MWP) solving is a long-standing challenging task in Natural Language Processing (NLP) and has attracted lots of attention recently~\cite{upadhyay2017annotating,upadhyay-etal-2016-learning,cass,dns,seq2et,trnn,sau-solver,recall-learn,generate-rank,NS-Solver,edge-enhanced}.
An automatic MWP solver should not only understand the problem's semantic information but also reason the grounded mathematical relationships implicit in the problem, so that it can transform natural language into solution expression.

\begin{figure}[t] 
	\centerline{\includegraphics[width=0.99\linewidth]{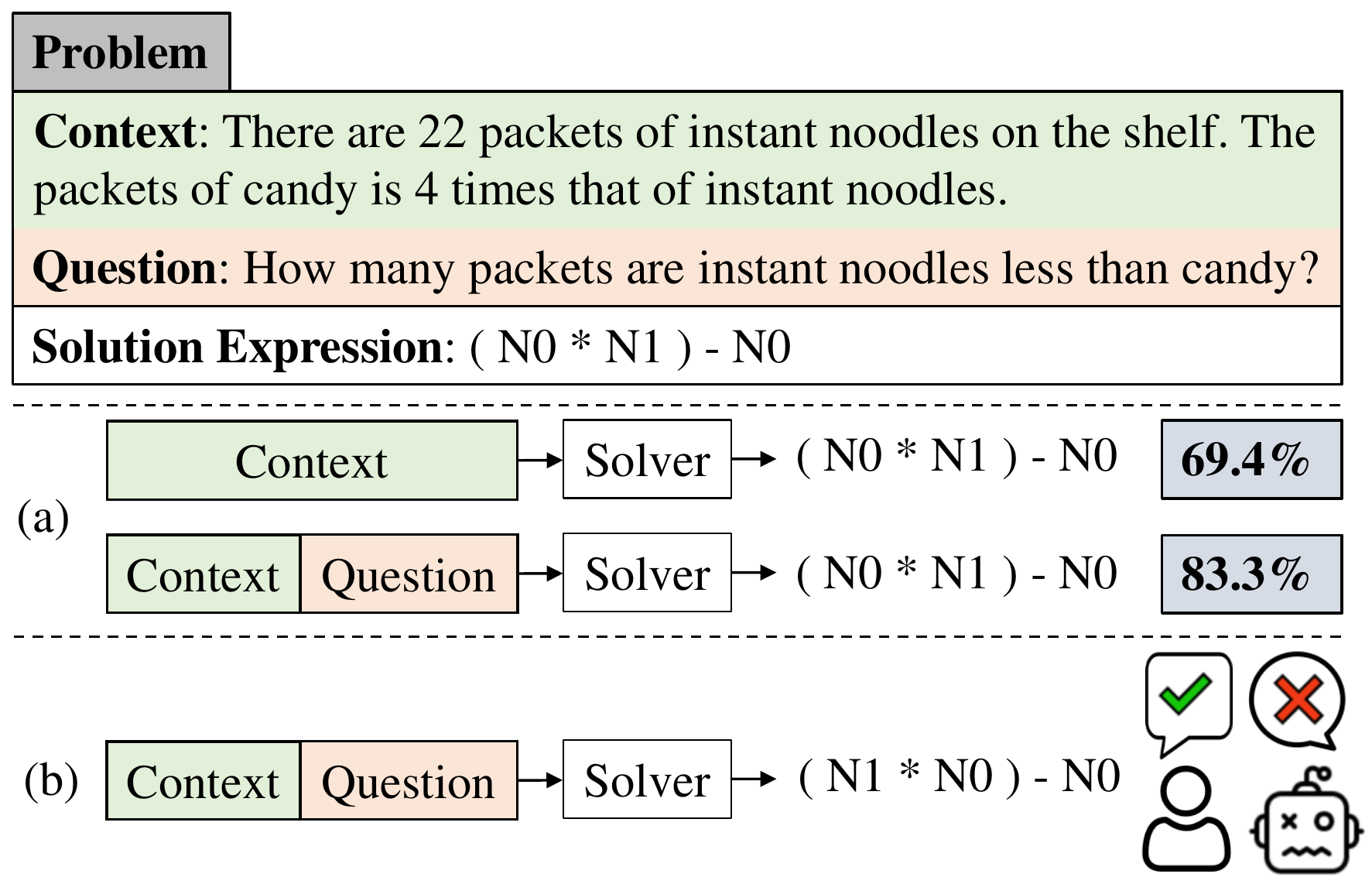}}
	\vspace{-2mm}
	\caption{Illustration of solving bias in MWP. A typical MWP problem can be divided into context and question. (a) shows that 69.4\% of the problems in Math23K can be answered by the solver (Bert2Tree) without looking at the question, verifying its severe data bias. (b) shows that the current training procedure ignores the equivalent expressions, indicating the possible learning bias.}
	\vspace{-5mm}
	\label{fig:example}
\end{figure}

More recently, deep learning methods~\cite{dns,seq2et,recall-learn,generate-rank,edge-enhanced} have made great progress in MWP solving and achieved impressive results on several popular benchmarks, such as Math23K~\cite{dns} and MAWPS~\cite{koncel2016mawps}. 
However, there exists some severe possible solving bias in these benchmarks, consisting of data bias and learning bias. Here, the \textbf{data bias} is introduced since the training dataset does not fully cover diverse questions for each problem narrative of all MWPs, leading to the situation that a solver only learns shallow heuristics rather than deep semantics for understanding problems. Besides, even the question of an MWP is deleted, a solver still can solve it correctly, as shown in Figure \ref{fig:example}(a). On the other hand, an MWP can be solved by multiple equivalent equations while current popular datasets only take one of the equivalent equations as the ground truth output for each sample, forcing the model to learn the labeled ground truth and ignore other equivalent equations which may be more suitable for a solver to learn, leading to \textbf{learning bias} during training. As shown in Figure \ref{fig:example}(b), if a solver may generate an answer-corrected expression that is different from ground-truth expression, it will be thought an error and the loss between the answer-corrected expression and the ground-truth expression will be back-propagated to the solver during training, leading to over-correct the solver. This learning bias makes it harder to learn to reason out answer-corrected expressions.

 
To mitigate the solver bias for pushing advanced models to learn underlying reasoning skills rather than solely matching nearest results, we first build a novel MWP dataset, UnbiasedMWP, to cover diverse questions for each problem narrative of all MWPs. It is constructed by varying the grounded expressions in our collected data and annotating them with corresponding new questions manually, thus mitigating data bias. Then, to mitigate the learning bias, we propose a Dynamic Target Selection (DTS) Strategy to dynamically select the most suitable target expression by applying the longest prefix match between the current model output and candidate equivalent equations obtained by applying commutative law during training. Our experimental result shows that our UnbiasedMWP has significantly fewer biases than its original data and other datasets, and the solvers equipped with our equivalent expression matching loss can achieve higher accuracy on multiple MWP benchmarks such as Math23K and our UnbiasedMWP. Our main contributions are in two folds:

\begin{itemize}
\setlength{\itemsep}{0pt}
\setlength{\parsep}{0pt}
\setlength{\parskip}{0pt}
\vspace{-3mm}
	\item We propose a large-scale data-unbiased dataset named UnbiasedMWP consisting of 10264 MWPs with diverse questions. The dataset is constructed by varying the grounded expressions and annotated corresponding questions. With this dataset, we can force a model to learn deep semantics rather than shallow heuristics for solving an MWP. 
	
	\item We propose a Dynamic Target Selection (DTS) Strategy to dynamically select a more suitable target expression, thus eliminating the learning bias caused by ignoring equivalent expressions during the training procedure. Experimental results demonstrate that the models trained with DTS achieve better performances on multiple benchmarks. Our DTS can improve the baseline model up to 1\%, 2.5\%, and 1.5\% on Math23K, UnbiasedMWP-Source, and UnbiasedMWP-All, respectively.   
\end{itemize}

\section{UnbiasedMWP dataset}
In this section, we introduce the construction procedure of our UnbiasedMWP dataset. Based on the newly-collected raw data, we design a pipeline for pre-processing and rewriting questions according to formula variations, which is strictly performed by the annotators to obtain unbiased data.  

\subsection{Data Collection and Pre-processing}
To collect UnbiasedMWP, we crawl 2907 examples from an online education website\footnote{\href{https://damolx.com/}{https://damolx.com/}}. During pre-processing, the number mapping~\cite{dns} is deployed to replace the numbers in solution expression with symbolic variables (e.g., $N0$, $N1$). Then, the workers are asked to split the problem text into two parts: context (a narrative implicated with numerical relationships) and question (a short text that requires the solution of a mathematical relationship). 

\subsection{Expression Variation}
\label{sec:EV}
As shown in Figure \ref{fig:example}, a neural network model can solve problems even without questions, this shows that a solver solves problems mainly by relying on shallow heuristics rather than deep semantic understanding. Besides, current popular and large-scale datasets do not fully cover any possible questions for the context in each MWP, which also results in data bias. To mitigate this issue, we annotate each narrative with various possible questions to construct an unbiased MWP benchmark by enumerating various expressions according to the number in the context, asking workers to design questions for each expression. If an expression can not be assigned with a suitable question, we remove it. 

To enumerate various possible expressions, we design three types of variation to create different expressions for each context:
\textbf{Variable assortment (Va) variations}: Selecting two variables from the context and combining them with the operators "$+, -, *, /$", such as $n_0 + n_1$, $n_0 - n_1$, etc. 
\textbf{Sub-expression (Sub) variations}: From the original expression, we choose all sub-expressions of it and change the operators to get new expressions. 
\textbf{Whole-expression (Whole) variations}: We get new expressions by changing the operators in the original expression. Besides, workers also can propose new expressions and annotate them.

Various expressions are first acquired by applying the variation processing. Then, we ask workers to write a practical question for each meaningful expression variation. For those meaningless expressions that can not be annotated with any practical question, we filtered out them. The details of data split and statistics are listed in the appendix.


\section{Dynamic Target Selection Strategy}

During the common MWP training procedure, only one expression is used as ground truth while the equivalent expressions are ignored. Consider the following case: the ground truth label is "$(N1*N0)-N0$" while the model output is "$(N0*N1)-N0$". Although they are mathematical equivalent, the model output is judged to be incorrect. Therefore, models are prone to be biased during training. To address this issue, we generate the equivalent expressions of the original ground truth expression and then select an equivalent expression matching the longest prefix with the current model output as target expression in the training procedure. 
  
\begin{figure}[t] 
	\centerline{\includegraphics[width=0.99\linewidth]{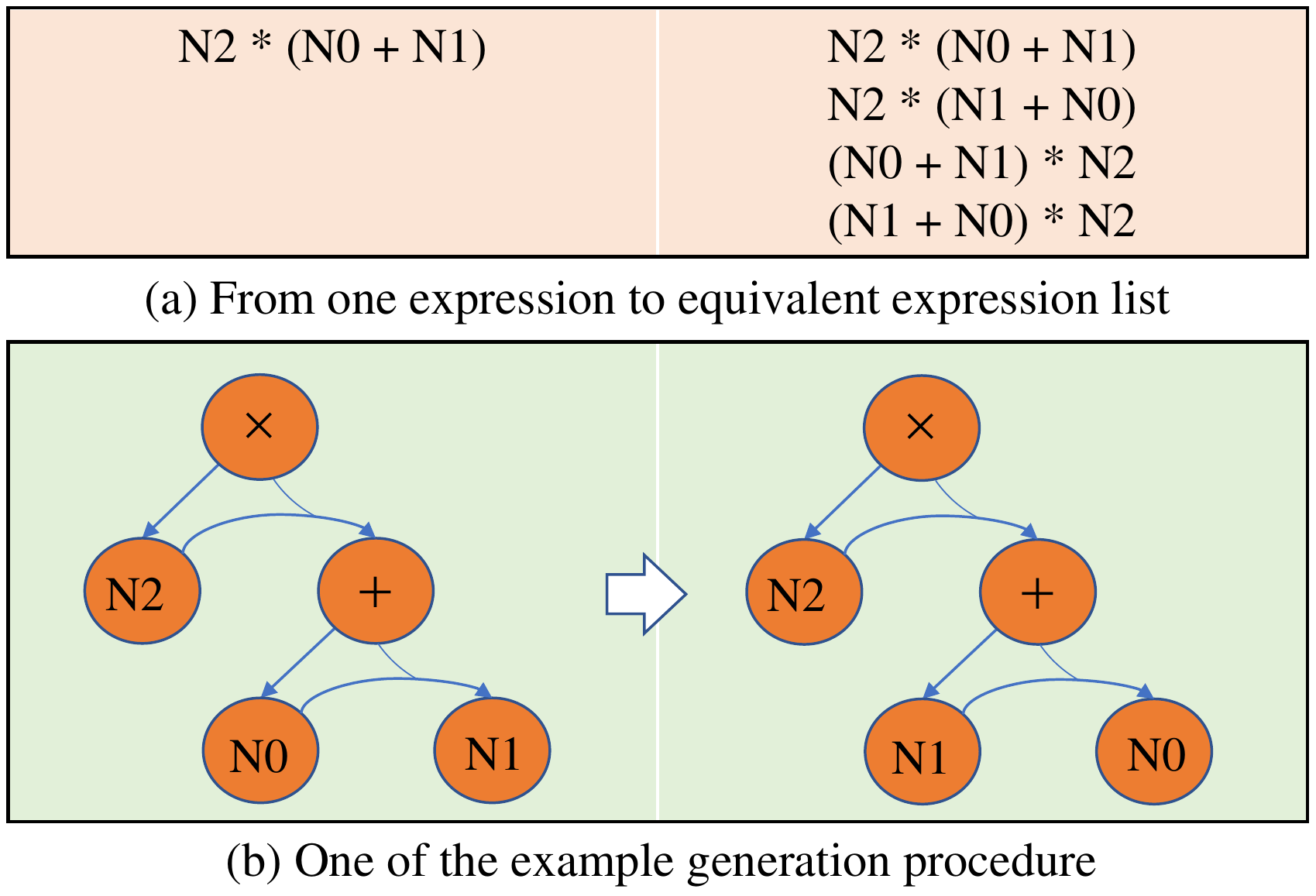}}
	\vspace{-3mm}
	\caption{Equivalent Expression Tree Generation. (a) shows the results of generation, (b) shows one of the generation examples.}
	\vspace{-5mm}
	\label{fig:dts}
\end{figure}

\subsection{Equivalent Expression Tree Generation}
To generate equivalent expressions, we consider swapping sub-expressions on the two sides of symmetric binary operators such as: $+$ and $\times$. Firstly, we construct an expression tree for each expression following \cite{seq2tree}. Then, we recursively examine each operator node from bottom to up and swap the left and right sub-trees of the node if it is a symmetric operator, and then we add the result new tree to a list. Finally, we iterate all the trees in the list into infix or prefix expressions to get multiple equivalent expressions. The generation procedure is illustrated in Algorithm \ref{alg:tree_variation}. 
An example of the generation is illustrated in Figure \ref{fig:dts} (b), we exchange the position of 'N0' and 'N1', and get a new equivalent expression. An example of generated equivalent expressions is shown in  Fig. \ref{fig:dts} (a).
\begin{algorithm}[t]
\small{
	\caption{Equivalent Expression Tree Generation}
	\label{alg:tree_variation}
	\textbf{Function}: \textit{Variation}($\mathrm{tree}$, $\mathrm{root}$, $\mathrm{equList}$)\\
	\KwIn{Expression tree: $ tree$; Root node of the input tree: $ root$;}
	\KwOut{Equivalent expression list:  $\mathrm{equList}$.}
	\BlankLine
	\If{\rm $ root$ is null}
	{
	return
	} 
	\textit{Variation}($\mathrm{tree}$, $\mathrm{root.left}$,  $\mathrm{equList}$)\\
    \textit{Variation}($\mathrm{tree}$, $\mathrm{root.right}$,  $\mathrm{equList}$)\\
    \If{\rm $\rm root.value$ is symmetric operator}
    {
        \textit{swap}($\mathrm{root.left}$, $\mathrm{root.right}$)\\
        $\mathrm{equList}$.append($\mathrm{tree}$)\\
        \textit{Variation}($\mathrm{tree}$, $\mathrm{root.left}$,  $\mathrm{equList}$)\\
        \textit{Variation}($\mathrm{tree}$, $\mathrm{root.right}$,  $\mathrm{equList}$)\\
        \textit{swap}($\mathrm{root.left}$, $\mathrm{root.right}$)
    }
    return 
}
\end{algorithm}

\subsection{Dynamic Target Selection (DTS)}
During the training procedure, the solver may generate the correct start part expression which matches the prefix of one of the equivalent expressions but not matches the prefix of the ground truth labeled in the dataset. If we still use the ground truth as the target to train the solver, this will lead to oversize error to correct the model prediction, leading to sub-optimal learning and learning bias. To mitigate this issue, we dynamically choose a new equivalent target expression as a training target that can match the current model output with the longest prefix. In this way, the loss will not be oversized so that we can make the solver easier to solve problems correctly. 




\section{Experiments}
\subsection{Experimental Setup}
\textbf{Datasets.} We conduct experiments on Math23K \cite{dns} dataset and our UnbiasedMWP dataset. We use \textbf{UnbiasedMWP-Source} to represent the initial collection of samples while using \textbf{UnbiasedMWP-All} to represent the initial collection of samples and their various variations. 

\noindent \textbf{Baselines.} We validate our UnbiasedMWP dataset and DTS training strategy with multiple models: \textbf{Math-EN}~\cite{seq2et}: a seq2seq model with equation normalization for reducing target space. \textbf{GROUPATT}~\cite{group-attn}: a solver borrowing the idea of multi-head attention from Transformer~\cite{transformer}. \textbf{GTS}~\cite{seq2tree}: a tree-structured neural network in a goal-driven manner to generate expression trees. \textbf{Graph2Tree}~\cite{graph2tree}: an enhanced GTS with quantity graph. \textbf{BERT2Tree}: a strong baseline we constructed by replacing RNN encoder with BERTEncoder\cite{chinese-bert-wwm} in GTS. More details can be referred to the appendix.

\subsection{Experimental Results}
\noindent \textbf{Bias Analysis on MWP Datasets}
We conduct similar experiments in \cite{svamwp} by removing question text on Math23K datasets and our collected UnbiasedMWP source data to show the solver mainly relied on shallow heuristics. As shown in Table \ref{tab:remove_q}, the experimental results on Math23K and UnbiasedMWP-Source show that all models still perform well even lack the question information. This suggests the patterns in the context have a strong correlation with the output expression, thus causing the model to learn bias in MWPs. We also conduct the same experiments on the UnbiasedMWP-All dataset. From Table \ref{tab:remove_q}, we can observe that the accuracies of the MWP without questions (Del\_q) are significantly lower on UnbiasedMWP-All than the other two datasets. This shows that our UnbiasedMWP can force the solver to solve an MWP with less bias.

\begin{table}[t]
\centering
\small
\setlength\tabcolsep{3pt} 
{
\begin{tabular}{lcccccc}
\toprule
 & \multicolumn{2}{c}{Math23K} & \multicolumn{2}{c}{\tabincell{c}{UnbiasedMWP \\ -Source}}&
 \multicolumn{2}{c}{\tabincell{c}{UnbiasedMWP \\ -All}}\\
 \cmidrule(lr){2-3} \cmidrule(lr){4-5} \cmidrule(lr){6-7}
 Methods & Ori & Del\_q & Ori & Del\_q & Ori & Del\_q\\
 \midrule
Math-EN & 68.4 & 55.2 & 62.0 & 42.5 & 52.5 & 20.4 \\ 
Group-Attn & 69.5 & 57.9 & 61.5 & 42.0 & 53.1 & 20.1 \\ 
GTS & 75.6 & 61.9 & 64.5 & 49.0 & 63.6 & 25.1 \\
Graph2Tree & 77.4 & 63.2 & 65.0 & 50.0 & 64.0 & 25.0 \\  
Bert2Tree & 83.3 & 69.4& 73.0 & 55.0 & 78.1 & 22.5 \\ 
\bottomrule	
\end{tabular}}
\vspace{-3mm}
\caption{Experimental results on Math23K, UnbiasedMWP-Source, and UnbiasedMWP-All. Ori indicates the original data and the Del\_q indicates data with the question removed.}
\label{tab:remove_q}
\vspace{-5mm}
\end{table}


\noindent \textbf{Robustness Analysis}
To further validate the advantages of our different variation data and how to improve a solver's robustness, we train two solvers on UnbiasedMWP-Source (Src) and UnbiasedMWP-All (All) and compare their performances on different test sets (Src, Src+Va, Src+Sub, Src+Whole, and All). From Table \ref{tab:vds}, we can observe that the solver trained with different variation data is more robust than the solver trained only with the initially collected samples on various test sets. This shows that our UnbiasedMWP can mitigate the learning bias of an MWP solver.

\begin{table}[t]
\centering
\large
\resizebox{0.99\linewidth}{!}{
\begin{tabular}{l|ccccc}
\toprule
\diagbox{Train}{Test} & Src & Src+Va & Src+Sub & Src+Whole & All \\ 
\midrule
Src & 73.0 & 37.3 & 49.7 & 53.1 & 34.9\\ 
All & \textbf{75.5} & \textbf{82.4} & \textbf{79.5} & \textbf{71.1} & \textbf{78.1}\\
\bottomrule
\end{tabular}}
\vspace{-3mm}
\caption{Comparison of results using different training and testing set. \textit{Va}, \textit{Sub}, and \textit{Whole} stand for the three variations mentioned in Section 2.2. \textit{All} denotes combining all three variations (\textit{Va} + \textit{Sub} + \textit{Whole}) on source (\textit{Src}) dataset.}
\label{tab:vds}
\vspace{-3mm}
\end{table}

\noindent \textbf{Analysis on DTS strategy} We conduct our DTS training strategy on Math23K and UnbiasedMWP. As shown in Table \ref{tab:empm}, our DTS training strategy helps several models achieve better performance. Especially, our DTS improves the accuracy of the Bert2Tree model from 83.3\% to 84.3\% on Math23K, from 73.0\% to 75.5\% on UnbiasedMWP-Source, and from 78.1\% to 79.6\% on UnbiasedMWP-All. In summary, the experimental results verify the validity of our DTS strategy.

\begin{table}[t]
\centering
\large
\renewcommand\arraystretch{1.1}
\setlength\tabcolsep{3pt}
\resizebox{0.99\linewidth}{!}{
\begin{tabular}{ccccc}
\toprule
Methods & DTS & Math23K & \tabincell{c}{UnbiasedMWP \\ -Source} & \tabincell{c}{UnbiasedMWP \\ -All} \\
 \midrule
GTS & \footnotesize{\XSolidBrush} & 75.6 & 64.5 & 63.6\\ 
\rowcolor{mycolor} GTS& \Checkmark & \textbf{76.4} & \textbf{65.5} & \textbf{63.7}\\ 
Graph2Tree & \footnotesize{\XSolidBrush} & 77.4 & 65.0 & 64.0 \\ 
\rowcolor{mycolor} Graph2Tree & \Checkmark & \textbf{77.8} & \textbf{65.5} & \textbf{64.6} \\ 
Bert2Tree & \footnotesize{\XSolidBrush} & 83.3 & 73.0 & 78.1\\  
\rowcolor{mycolor} Bert2Tree & \Checkmark & \textbf{84.3} & \textbf{75.5} & \textbf{79.6}\\ 
\bottomrule
\end{tabular}}
\vspace{-2mm}
\caption{Comparison of experimental results with or without DTS of GTS-based \cite{seq2tree}  model.}
\label{tab:empm}
\vspace{-5mm}
\end{table}
\section{Conclusion}
\vspace{-2mm}
In this paper, we revisit the solving bias in MWP. To mitigate the data bias caused by lacking question diversity, we construct a data set called UnbiasedMWP by variating the expressions in new-collected data. The experimental results illustrate that the solver trained on UnbiasedMWP is more robust than on our collected data. To mitigate the learning bias caused by loss overcorrect with taking only one ground-truth, we proposed a strategy to generate the equivalent expressions and select the longest prefix with the current model output during training, called Dynamic Target Selection (DTS). Experimental results show that our DTS helps several models achieve better performance.

\section*{Acknowledgements}
This work was supported in part by National Key R\&D Program of China under Grant No. 2020AAA0109700, National Natural Science Foundation of China (NSFC) under Grant No.U19A2073 and No.61976233, Guangdong Province Basic and Applied Basic Research (Regional Joint Fund-Key) Grant No.2019B1515120039, National Natural Science Foundation of Guangdong Province (Grant No.2022A1515011835), China Postdoctoral Science Foundation funded project (Grant No. 2021M703687), Shenzhen Fundamental Research Program (Project No. RCYX20200714114642083, No. JCYJ20190807154211365), CAAI-Huawei MindSpore Open Fund. We also thank MindSpore for the partial support of this work, which is a new deep learning
computing framework\footnote{https://www.mindspore.cn/}.

\bibliography{anthology,custom}
\bibliographystyle{acl_natbib}

\appendix

\section{Appendix}
\label{sec:appendix}

\subsection{Data Split}

To ensure that the model does not see the context from the /testing set during training, We first split the training, validation, and testing set on our newly collected source dataset. Then we further apply the expression variation (mentioned in Section \ref{sec:EV}) to expand the data on different subsets. The size of the split of our collected data and variation data is shown in Table \ref{tab:split}.

\begin{table}[htbp]
\centering
\small
{
\begin{tabular}{l|cc}
\toprule
Split & \tabincell{c}{UnbiasedMWP \\ -Source} & \tabincell{c}{UnbiasedMWP \\ -All} \\ 
 \midrule
Train & 2507 & 8895\\ 
Validation & 200 & 684\\ 
Test & 200 & 685\\ 
\bottomrule
\end{tabular}}
\caption{Size of UnbiasedMWP data split.}
\label{tab:split}
\vspace{-3mm}
\end{table}

\subsection{Examples of data variation}
\begin{figure}[htbp] 
	\centerline{\includegraphics[width=0.99\linewidth]{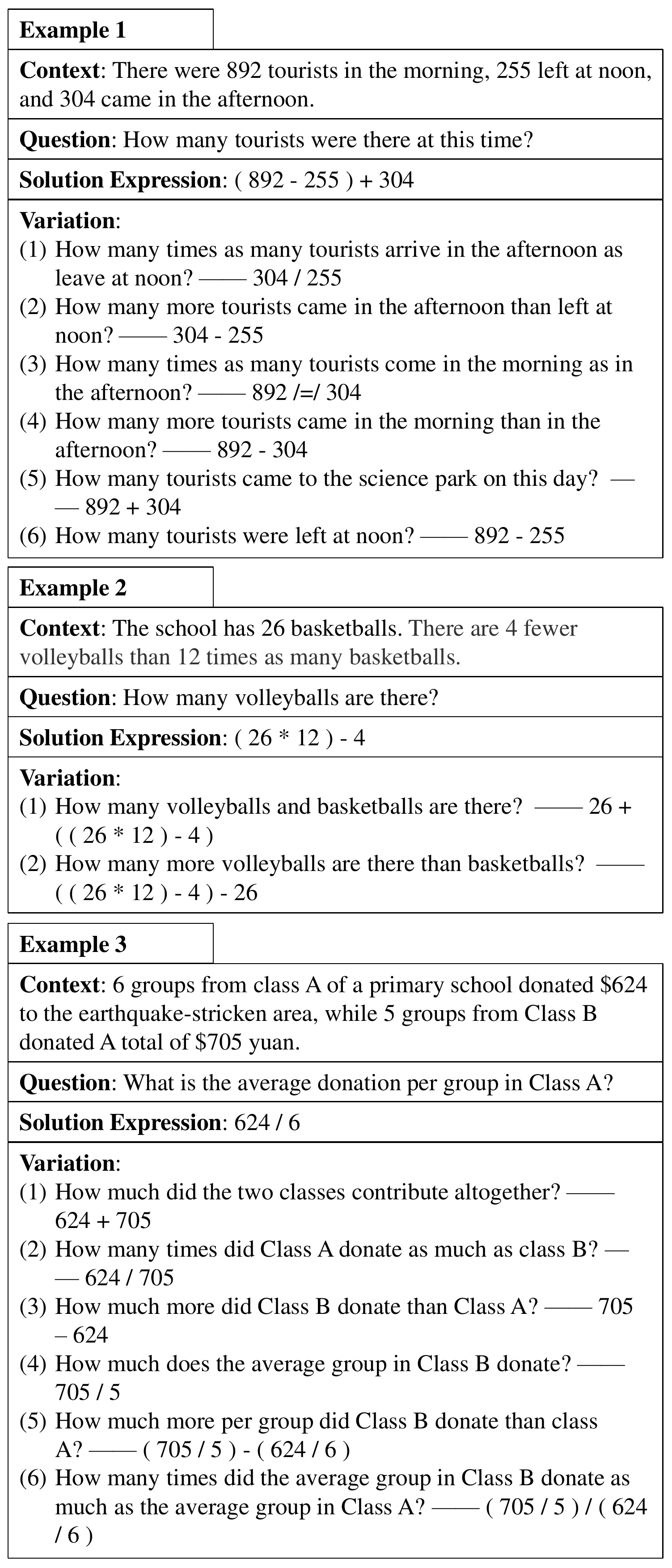}}
	\vspace{-3mm}
	\caption{Some examples of our data variation.}
	\vspace{-3mm}
	\label{fig:variation_example}
\end{figure}

Figure \ref{fig:variation_example} shows some examples of our data variation.

\subsection{Data statistic}
We analyze the proportions of data of different prefix expression lengths in UnbiasedMWP dataset and the result is shown in Table \ref{tab:exp_length}. We analyze our UnbiasedMWP to count the size of different variation data, the statistical result is shown in Table \ref{tab:variation}. Note that the count of All data is not equal to the sum of the above rows in the table, because there will be some overlap between the variation data obtained in the three data variation methods mentioned in Section \ref{sec:EV}.

\begin{table}[htbp]
\centering
\small
\setlength\tabcolsep{3.7pt} 
{
\begin{tabular}{l|cccccc}
\toprule
Expression Length & 3 & 5 & 7 & 9 & 11 & >11 \\ 
\midrule
Count & 6357 & 2560 & 1011 & 215 & 90 & 31 \\
\bottomrule
\end{tabular}}
\caption{Statistics analyse on prefix expression length.}
\label{tab:exp_length}
\vspace{-3mm}
\end{table}

\begin{table}[htbp]
\centering
\small
{
\begin{tabular}{l|c}
\toprule
 & Count \\ 
 \midrule
Source & 2907 \\ 
Variable assortment & 5083 \\ 
Sub-expression & 2843 \\ 
Whole expression  & 2205 \\ 
All data & 10264 \\ 
\bottomrule
\end{tabular}}
\caption{Statistics analyse on variation data.}
\label{tab:variation}
\vspace{-3mm}
\end{table}

We also analyze the accuracy of data of different prefix expression lengths for Bert2Tree model shown in Table \ref{tab:result_length}. Experimental results show that the longer the expression, the lower the accuracy.

\begin{table}[htbp]
\centering
\small
\setlength\tabcolsep{3.7pt} 
{
\begin{tabular}{l|cccc}
\toprule
Expression Length & 3 & 5 & 7 & >= 9\\ 
\midrule
Count & 85.3 & 74.3 & 60.30 & 26.5\\
\bottomrule
\end{tabular}}
\caption{Performance of Bert2Tree on different prefix expression length of UnbiasedMWP-All.}
\label{tab:result_length}
\vspace{-3mm}
\end{table}

\subsection{Implementation details}
Pytorch\footnote{http://pytorch.org} is used to implement our our MWP solver on Linux with NVIDIA RTX1080Ti GPU card. Our \textbf{Bert2Tree} model is constructed by replacing the encoder in \textbf{GTS} model with the Chinese Bert\cite{chinese-bert-wwm}. The learning rate is set as $5e^{-5}$ and $1e^{-3}$ for Bert encoder and tree-decoder respectively. Adam is set as the optimizer of \textbf{Bert2Tree} while $\beta_1$ = 0.9, $\beta_2$ =0.999, and $\epsilon$ = $1e^{-8}$. The batch size is $32$. Dropout weight is set as $0.5$ with weight decay $1e^{-5}$. For the other four models, \textbf{Math-EN}, \textbf{Group-Attn}, \textbf{GTS} and \textbf{Graph2Tree}, we follow their original parameter settings in \cite{SMART}. Since the data pre-processing code in \textbf{Graph2Tree} is not open, we do not evaluate this model on our own data. 

In the experiments, we train Bert2Tree for 100 epochs on Math23K while 50 epochs on our UnbiasedMWP-Source and UnbiasedMWP-All data, because Math23K is a larger benchmark dataset whch contains 23K samples. For the Del\_q experiments, We intercept the last sentence (question) by detecting punctuation marks in Math23K which may cause some very small errors but does not affect the overall results of the experiment. For our UnbiasedMWP dataset, we directly use the context to do the Del\_q experiment.

\subsection{Related Work}
\textbf{Math Word Problem Solving} 

In recent years, deep learning models especially Seq2Seq models\cite{dns, group-attn, seq2et, seq2tree, graph2tree, NS-Solver, generate-rank, edge-enhanced}, have made great progress in MWPs by learning to translate problem text in natural language into mathematical solution expression. \cite{dns} is the first to apply deep learning in MWPs and propose a widely used dataset called Math23K. \cite{group-attn} propose a group attention mechanism to extract multi-dimensional features. \cite{seq2tree} propose a tree decoder to decode expression as prefix order. Based on \cite{seq2tree}, \cite{graph2tree} improve the encoder embedding by fusing a graph encoder's output. \cite{NS-Solver} propose a framework by applying multiple auxiliary tasks to improve the problem embedding and the ability to predict commonsense constants. \cite{generate-rank} devise a new ranking task for MWP and propose the Generate \& Rank, a multi-task framework based on a generative pre-trained language model. \cite{edge-enhanced} propose a novel Edge-Enhanced Hierarchical Graph-to-Tree model (EEH-G2T), in which the math word problems are represented as edge-labeled graphs.
 
\paragraph{Challenging Datasets and Adversarial Examples of MWP}

More challenging datasets in MWP are proposed in recent years, Ape210K \cite{ape210k} provides a large-scale benchmark for evaluating MWP solvers,
HMWP \cite{sau-solver} is a Chinese MWP benchmark including examples with multiple-unknown variables requiring non-linear equations to solve.

Although solvers have achieved impressive performance on these datasets, the robustness of the solvers is questioned in \cite{ad-mwp}. Besides, \cite{svamwp} also points out that MWP solvers rely on shallow heuristics to achieve high performance and propose SVAMWP dataset. SVAMWP is more reliable and robust for measuring the performance of MWP solvers, because it raises the requirement for the model's sensitivity to question text through applying variations over word problems. Unlike SVAMWP, our variations are applied to expressions to get different expression-question pairs.

\subsection{Analysis on Effects of PLM}
We conduct experiments on our UnbiasedMWP dataset as shown in Table \ref{tab:remove_q}. In the experiments, for models without PLMs such as \textbf{Math-EN} \cite{dns}, \textbf{GROUPATT} \cite{group-attn}, \textbf{GTS} \cite{seq2tree}, \textbf{Graph2Tree} \cite{graph2tree}, they perform worse on UnbiaseMWP-All dataset than on UnbiasedMWP-Source dataset, whereas for \textbf{Bert2Tree} model with Bert \cite{chinese-bert-wwm}, it performs significantly better on UnbiasedMWP-All dataset. This shows that the UnbiasedMWP-ALL dataset with diverse questions is more likely to confuse the model, because the context of the sample remains unchanged and only changes the question. However, Bert2tree can better distinguish the difference between diverse questions through the pre-trained language model.

\end{document}